\documentclass{SCIS2021}

\usepackage[english]{babel}
\usepackage{amsthm}
\usepackage[colorlinks, linkcolor=black]{hyperref}
\newtheoremstyle{exampstyle}
{0.0em} 
{0.0em} 
{} 
{1em} 
{\bfseries} 
{.} 
{1em} 
{} 
\usepackage{balance}
\theoremstyle{exampstyle}

\usepackage{CJKutf8}

\usepackage{multirow}
\usepackage{multicol}
\usepackage{booktabs}
\usepackage{adjustbox}


\begin{document}
	\ArticleType{RESEARCH PAPER}
	\Year{2021}
	\Month{}
	\Vol{}
	\No{}
	\DOI{}
	\ArtNo{}
	\ReceiveDate{}
	\ReviseDate{}
	\AcceptDate{}
	\OnlineDate{}
	
	\title{Integrated Structural Prompt Learning for Vision-Language Models}
	
	\author[1, 2]{Jiahui Wang}{}
        \author[1, 2]{Qin Xu}{{xuqin@ahu.edu.cn}}
        \author[1, 3]{Bo Jiang}{}
        \author[1, 2]{Bin Luo}{}
	\AuthorMark{Author A}
	
		\AuthorCitation{Jiahui Wang, Author B, Author C, et al}
	
	
\address[1]{School of Computer Science and Technology, Anhui University, Hefei {\rm 230601}, China}
\address[2]{The Key Laboratory of Intelligent Computing and Signal Processing of Ministry of Education}
\address[3]{Information Materials and Intelligent Sensing Laboratory of Anhui Province}

	\abstract{Prompt learning methods have significantly extended the transferability of pre-trained Vision-Language Models (VLMs) like CLIP for various downstream tasks. These methods adopt handcraft templates or learnable vectors to provide text or image instructions in fine-tuning VLMs. However, most existing works ignore the structural relationships between learnable prompts and tokens within and between modalities. Moreover, balancing the performance of base and new classes remains a significant challenge. In this paper, we propose an Integrated Structural Prompt (ISP) for VLMs to enhance the interaction of information representations between the text and image branches. ISP introduces self-structural and cross-structural prompt modules to model the structural relationships between learnable prompts and frozen tokens within and across modalities. This enables efficient information transfer while preserving feature stability. Additionally, we propose a sample probing module that dynamically adjusts loss coefficients based on sample difficulty, preventing the mode from overfitting to simple samples and improving generalization ability to new classes. Extensive experiments on three widely used settings: base-to-new generalization, cross-dataset evaluation, and domain generalization demonstrate that the proposed ISP achieves competitive performance against state-of-the-art methods. }
	
	\maketitle

\section{Introduction}
The deep learning models that are pre-trained on large-scale datasets \cite{imagenet, places} are widely used in various downstream tasks, such as image classification \cite{resnet, vit, mpsa}, object detection \cite{rcnn, yolo}, and semantic segmentation \cite{semantic1, sam}.
With the recent application of image-text datasets, large-scale vision-language models (VLM) \cite{lit, filip, align, flamingo, clip} have been widely used. For example, CLIP \cite{clip} uses 400 million image-text pairs to pre-train the vision-language models and demonstrates strong potential in multi-modal tasks. However, the huge parameters of VLMs make it very difficult to fine-tune and transfer to other tasks with limited labeled data. 

Prompt learning has been widely adopted to migrate VLM efficiently. Early works \cite{handprompt, promptnlp} usually design handcrafted templates for text token sequences to transfer VLM to different tasks. For example, CLIP \cite{clip} first pre-trained the VLMs with image-text pairs and then adopted handcraft templates to build text prompts to adapt to the image classification task. However, pre-designed prompt words are not flexible enough for further fine-tuning. CoOp \cite{coop} proposed to use learnable prompts instead of manually designed prompt words for the class names and freeze the parameters of the CLIP encoder to better generalize for the downstream tasks. KgCoOp \cite{kgcoop} proposed to regularize the features extracted by learnable prompts with the frozen text features extracted by handicraft prompts to reduce the forgetting of original knowledge. HPT \cite{hpt} suggested the use of hierarchical descriptions to supplement text features with linguistic knowledge. However, the above methods based on prompt engineering usually learn prompts separately within each modality, hindering information relevance modeling. Thus, Maple \cite{maple} adds learnable prompts to both text and visual token sequences and transfers information from the text branch to the visual branch through prompts. COMMA \cite{comma} proposed the fusion of the text prompts and visual prompts of previous layers into post layers to pass messages between modalities. MMA \cite{mma} proposed to use adapters to fine-tune the network instead of the design of prompts and add a shared linear projection between the two modalities to transfer information.


However, we found that the existing work still faces several issues: 1) the structural relevance of the learnable prompts within and between modalities is not effectively utilized. The learnable prompts represent domain-related but class-independent knowledge, while the frozen image tokens bring more discriminative class-related knowledge.
During fine-tuning, it is difficult for the frozen encoder layer to construct the relationship between the two types of tokens, which hinders the interaction between domain information and sample information. 2) There is a significant difference in the feature distribution between text and image features, and simply conducting feature interaction between multi-modal features may lead to network instability. 3) Due to the limited number of labeled samples, previous methods often overfit simpler samples from the base classes and demonstrate suboptimal generalization abilities to new classes.

To solve the above problems, we propose integrated structural prompts for visual language models. Specifically, in order to make full use of the structural relevance between prompts and tokens, our proposed self-structural prompt constructs correlations between prompt and text/visual tokens by performing information transfer by cross attention. The optimized prompts fuse the features of the samples contained in frozen tokens while retaining the domain-related information in learnable prompts. To highlight important information and improve the efficiency of information transfer, we only use the tokens with higher responses to construct the correlation matrix. Then, to effectively transfer information between modalities and keep feature stability, we design the cross-structural prompt module. By constructing an affinity graph between the prompt of one modality and the tokens of another modality, the internal relation representation of the prompts is constructed. Then, graph convolution is used to transfer structural information between modalities to refine the prompts. Finally, to solve the problem of overfitting to simple samples, we propose the sample probing that defines the difficulty of each sample according to the difference between predicted probability after the prompt and the prediction results of zero-shot CLIP and sets different loss coefficients. Extensive experiments under widely used settings, including base-new generalization, cross-dataset evaluation, and domain generalization settings, demonstrate that our method achieves leading results on most datasets. These results display a strong generalization ability to downstream image classification tasks.
In general, our contributions are summarized as follows:
\begin{itemize}
    \item Our proposed integrated structural prompt fully utilizes the structural relationship between learnable prompts and frozen tokens to transfer information within and between modalities.
    \item The proposed self-structural prompt and cross-structure prompt to refine the learnable prompts while keeping the stability of features.
    \item We design the sample probing that applies different loss coefficients to each sample, thereby reducing the weights of the model to simple samples to improve the generalization ability.
\end{itemize}


\section{Related Works}
\label{sec:related}
\subsection{Vision-Language Models}

The ability to learn joint multi-modal representations has been significantly advanced by Visual Language Models (VLMs) like CLIP \cite{clip}, ALIGN \cite{align}, FLIP \cite{filip}, and BLIP \cite{blip}. These models leverage contrastive learning, that maximizes similarity for matched image-text pairs while minimizing it for unmatched ones and benefit greatly from pre-training on large-scale datasets \cite{laion5b}, demonstrating strong performance on downstream tasks. However, a major limitation arises from their large size and dependence on extensive training data, making deployment difficult in specialized environments, especially when labeled data is scarce (few-shot settings). This bottleneck has spurred investigation into methods for efficiently adapting pre-trained VLMs, as seen in recent studies \cite{clipadapter, tipadapter, promptdet, regionclip, opendet, clipseg, extractclip} focusing on improving performance in few-shot and zero-shot contexts for various applications.

\subsection{Prompt Learning in VLMs}
Inspired by efficient transfer learning \cite{transnlp, lora} in natural language processing (NLP), adapting VLM to specific tasks with minimal training parameters attracts more attention.  Early research in NLP \cite{promptnlp} shows that adding task-specific prompts to a pre-trained model can significantly improve its performance on downstream tasks. The key idea of prompt-based methods in VLMs is to use hand-crafted templates \cite{clip} or learnable vectors \cite{coop} to the input features to fit the pre-trained model. For example, CoOp \cite{coop} introduced learnable vectors as prompts to replace predefined text prompts in the language branch of CLIP to improve its performance in few-shot tasks. CoCoOp \cite{cocoop} further extends this idea by introducing image conditional text prompts, allowing the model to adapt specific image instances more flexibly. Other methods, such as ProGrad \cite{prograd} and KgCoOp \cite{kgcoop}, try to optimize the alignment between the output of the text encoder and the pre-trained CLIP model to retain beneficial knowledge during fine-tuning. PLOT \cite{plot} learns multiple local prompts by introducing optimal transport distance to match local visual features with text features. SEP \cite{sep} proposes to further enhance the learnable prompts using frozen visual text markers to improve the generalization ability of the prompts. While these works have demonstrated the effectiveness of learnable prompts, they have typically focused on the separate optimization of prompts within each modality (text or visual) without fully exploiting the interaction between the two modalities. 

\subsection{Multi-modal Prompt Learning}
Recent studies have explored multi-modal prompt learning to address these limitations, i.e., simultaneously learning prompts for the textual and visual branches of VLMs. MaPLe \cite{maple} proposed to learn prompts in the visual branch and the language branch simultaneously and transfer prompts from the text branch into the visual branch to achieve the sharing of information between modalities. This approach still lacks the two-way interaction between prompts and does not fully exploit the structural relationship between the two modalities. COMMA \cite{comma} proposes to aggregate two branch prompts to generate prompts for the visual branch in subsequent blocks, thereby obtaining multi-modal useful information and performing feature alignment. However, this method cannot achieve bi-directional interaction due to the feature differences between modalities. MMA \cite{mma} proposes to use adapters for the visual and text branches and add a linear projection layer shared by the two branches in the middle of the adapter to achieve feature interaction. 

\subsection{Model Regularization}
The strong fitting ability of prompt learning often causes the model to forget the original general knowledge, thereby reducing the generalization ability of the model on the unseen new classes. To solve this problem, LAMM \cite{lamm} proposed the use of a hierarchical regularization method to regularize the model on learnable prompts, text features, and prediction probabilities. COMMA \cite{comma} proposed to regularize the text token features after the prompt with the text features of the frozen CLIP model in each block to retain the general knowledge of pre-trained CLIP for the model. MetaPrompt \cite{metaprompt} proposed to use an asymmetric contrastive learning strategy to prevent the learned prompt vector from overfitting. DePT \cite{dept} discovered the relationship between channel feature distribution and base-new generalization ability and proposed to add channel transformation adapters to the base class dataset to retain CLIP general knowledge. AMU \cite{amutuning} proposed to fuse the prediction results of the zero-shot CLIP model according to the uncertainty of the prediction probability to improve the robustness of the model. However, existing work does not consider the relationship between the difficulty of each sample and the loss weights of each sample. 

\section{Method}
\label{sec:method}

\subsection{Preliminary of CLIP}
\label{sec:clip}


The CLIP architecture consists of two core components: a visual encoder $V$ and a text encoder $T$. This dual-encoder structure is designed to learn a shared embedding space for images and text. Pre-training occurs on vast datasets of image-text pairs using a contrastive loss, which effectively teaches the model to maximize the similarity between corresponding image and text representations while minimizing it for non-corresponding pairs. This process endows CLIP with a robust understanding of visual-semantic connections and excellent generalization capabilities. Visual Encoder $V$ takes an image $I$ and extracts the visual feature $x' \in \mathbb{R}^{d_v}$ as:
\begin{align}
    X_0' =& \mathrm{Embed}_V(I)\in \mathbb{R}^{M\times d_v}\\
[c_i, X_i'] =& V_i([c_{i-1}, X_{i-1}'])\quad i = 1, 2, \dots, L\\
x' =& \mathrm{FC}_V(c_L')
\end{align}
where $\mathrm{Eemb}_V$ and $\mathrm{FC}_V$ represent the embedding layer and feature projection layer of the visual encoder, respectively, $M$ denotes the length of the visual token sequence, and $L$ is the number of layers of the encoder. The learnable class token $c_0$ is concatenated to the visual token sequence and used for the final classification, and  $x'$ represents the visual feature. 
The text encoder $T$ takes templates descriptions $\mathcal{T}$ for each class (e.g., "a photo of a \texttt{[class]}"). These templates are tokenized and embedded into $W'_0$. These embeddings are processed by $L$ transformer layers ($T_i$) and projected by $\mathrm{FC}_T$ to obtain the text features $w'$ for all classes as
\begin{align}
    W'_0 =& \mathrm{Embed}_T(T)\in \mathbb{R}^{C\times N\times d_t}\\
W'_i =& T_i(W'_{i-1})\quad i =  1, 2, \dots, L\\
w' =& \mathrm{FC}_T(W_L')
\end{align}
where $\mathrm{Eemb}_T$ and $\mathrm{FC}_T$ represent the embedding layer and feature projection layer of the text encoder, $N$ represents the length of the text token sequence, and $w$ represents the extracted text features. The text and visual features are calculated by the cosine similarity function for prediction:
\begin{equation}
    p(y|x') = \frac{\mathrm{exp}(\cos(x', w_i')/\tau)}{\sum_{j=1}^k\mathrm{exp}(\cos(x', w_j')/\tau)}
\end{equation}
where $\tau$ is the temperature hyperparameter, $\cos$ represents the cosine similarity, and $y$ represents the ground truth label of image $I$.


\begin{figure}[t]
    \centering
    \includegraphics[width=0.75\linewidth]{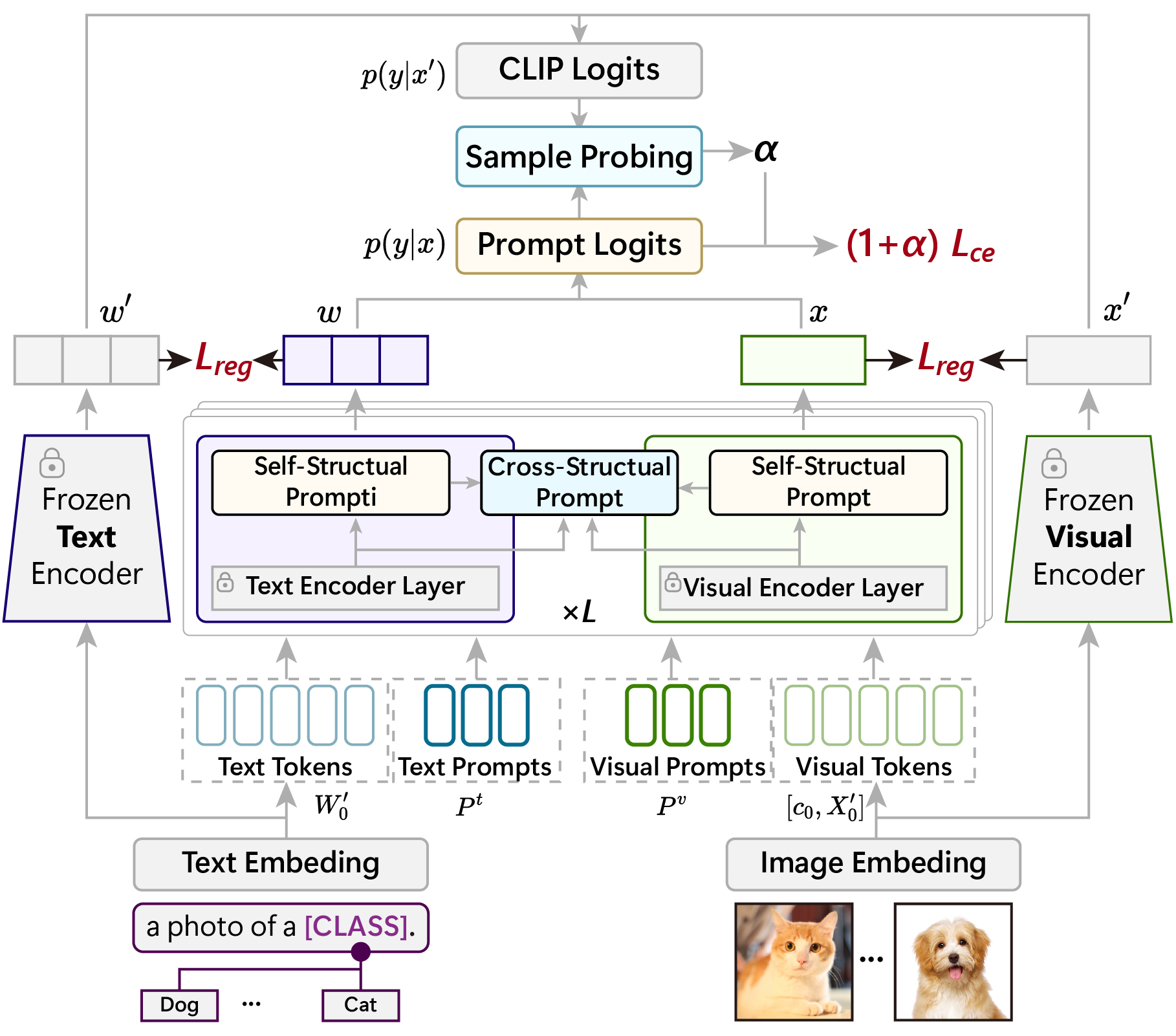}
    \caption{Overview of the proposed method. 
    Firstly, the text tokens and visual tokens are concatenated with learnable prompts respectively. Then the text and image features are extracted by the frozen encoder and the encoder adds the integrated structural prompt. Finally, the sample probing adds an adaptive loss coefficient by the sample difficulty which is calculated by the prediction probability of frozen clip and prompted encoder.
    }
    \label{fig:overview}
\end{figure}

\subsection{Overview}
\label{sec:overview}

\begin{figure*}[t]
    \centering
    \includegraphics[width=\linewidth]{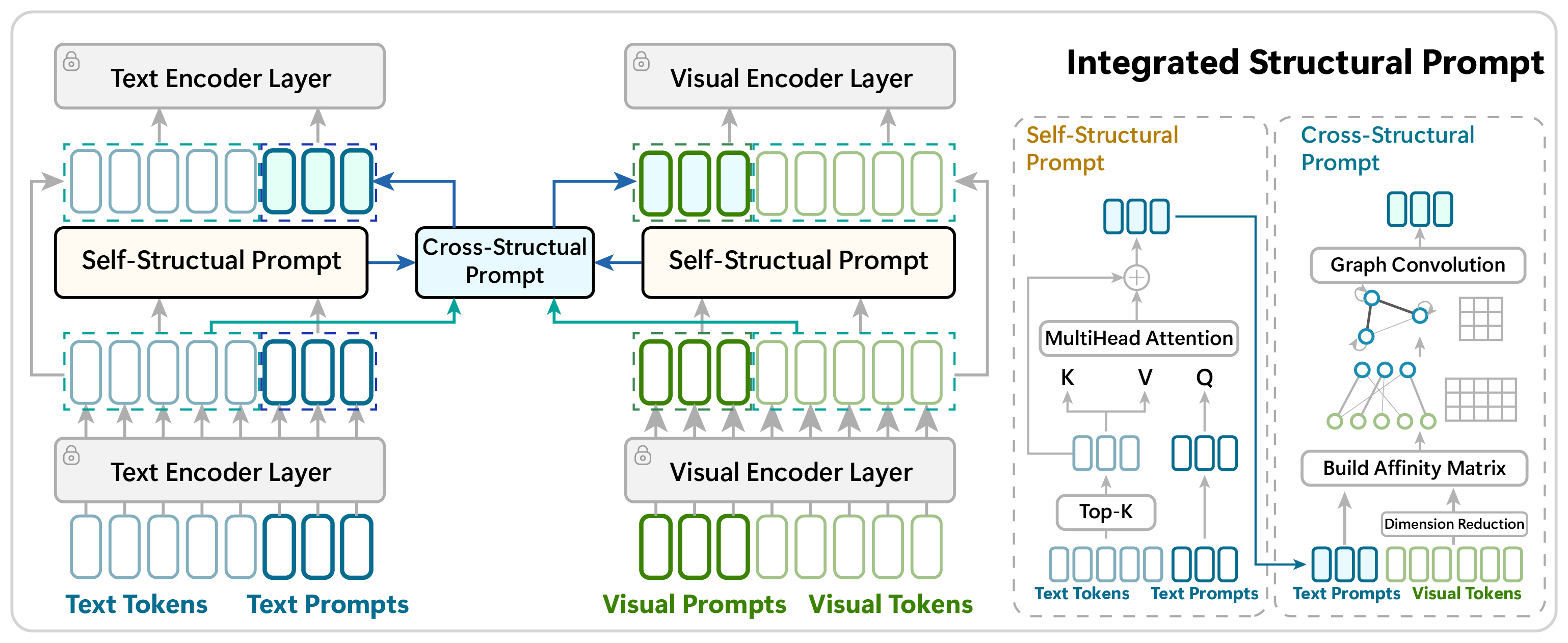}
    \caption{The diagram of the proposed integrated structural prompt (Left) and the detailed implementation of self-structural prompt and cross-structural prompt (Right). 
    }
    \label{fig:isp}
\end{figure*}

The overall diagram is demonstrated in Figure \ref{fig:overview}.
Firstly, the learnable prompts are embedded into visual and text sequences, represented by $X_0 = [x_0', P^v]$, and $W_0=\{w_0^j\}^C_{j=1}$, $w_0 =[w'_0, P^t]$, where $P^v\in \mathbb{R}^{L_v\times d_v}$, $P^t\in \mathbb{R}^{L_t\times d_t}$ are the learnable prompts, $L_v$ and $L_t$ denote the number of vision and text prompts
Then, the visual and language sequences are input into the frozen encoder of each layer to extract features and then input into our proposed self-structured prompt models to construct the structural relationship between tokens and prompts. Next, the output of the prompt by self-structured prompting is fed to the cross-structured prompt to optimize the prompt feature representation by utilizing the constructed structural relationship between text and visual tokens. The optimized text and visual prompts are concatenated with the token sequence as the input of the next layer. Finally, the proposed sample probing module dynamically adjusts the loss weight of each sample to reduce the overfitting phenomenon during training. The overall diagram can be referred as Figure \ref{fig:isp}.

\subsection{Self-Structural Prompt}
\label{sec:ssp}
To fully integrate the relationship between the prompts and the tokens, we explicitly construct the structural relationship between the prompts and the tokens. We fuse the class-related knowledge of the tokens with the prompt features by the proposed self-structural prompt. The optimized prompts can fuse the features of the samples while retaining the domain-related information to improve the performance of the network in both base classes and new classes.

Specifically, taking the visual branch as an example, for the visual tokens of the $l$-th layer in the network, we denote it as $X_l=[c_l, x_l, P^v]\in \mathbb{R}^{(M+L_v+1)\times d_v}$. If we directly construct the relationship between the prompt and all of the tokens, it will bring a huge computation cost. At the same time, referring to \cite{sep}, we believe that only a few tokens are discriminative for classification, whether in visual or textual modality. To incorporate more discriminative features for the prompts, we first select $L_v$ visual tokens, which are the same as the number of prompts. We calculate the sum of the squares of the values of each token in the channel dimension as the response value and select $L_t$ tokens $\tilde{x}\in \mathbb{R}^{L_t\times d_v}$ with the largest response values, as follows:
\begin{equation}
    \tilde{x} =  \mathrm{TopK}(x_l, \sum_{i=1}^{d_v} x_{li}^2)
\end{equation}
We then use the similarity between the learnable prompts and the selected tokens as the structural relationship and aggregate features based on their correlation. If we use dot production attention as the similarity function, the above process can be expressed in the form of cross attention, that is:
\begin{equation}
    \widetilde{P^v_l} = \mathrm{Softmax}(\frac{P^v_l\tilde{x}^\top}{\sqrt{d_v}})\tilde{x}
\end{equation}
In the implementation, we add the projection parameters of Query, key, and Value to Eq 5, respectively, and add a 2-layer MLP to map the aggregated prompts, which can be expressed as:
\begin{align}
    \widetilde{P^{v}_l}’ &= \mathrm{CrossAttn}(\mathrm{LN}(P^v_l), \mathrm{LN}(\tilde{x}), \mathrm{LN}(\tilde{x}))+\tilde{x}\\
\widetilde{P^{v}_l} &= \mathrm{MLP}(\mathrm{LN}(\widetilde{P^{v}_l}’))+\widetilde{P^{v}_l}’
\end{align}
where $\mathrm{LN}(\cdot)$ represents layer normalization. We adopt the same operations for prompts in the text encoder branch. By aggregating features through the structural relationship between prompts and tokens, we can effectively integrate domain-related knowledge and discriminative class knowledge, thereby effectively improving the generalization ability of the network for new classes.

\subsection{Cross-Structural Prompt}
\label{sec:csp}
Existing multi-modal prompt learning methods usually only use learnable prompts to transfer information from text to visual modality in a unidirectional manner and cannot fully interact with information between modalities. 
We believe there are two main difficulties in existing information transfer: 1) the dimensions of vision and text are usually not uniform, and 2) the distribution of visual and text features is quite different. 
To optimize prompts by the structural relationship between modalities, we propose cross-structural prompts. 
By explicitly constructing the relationship between modalities and aggregating information in the prompts within the modality, the model enhances the information interaction between modalities while keeping the stability of the features to better adapt to various types of downstream tasks.

Specifically, we first reduce the dimension of the visual tokens that have higher dimensions. We use parameter-free discrete cosine transform and retain low-frequency channels with higher information density to achieve dimensionality reduction with less information loss. The visual feature $\overline{X'_l}\in \mathbb{R}^{(M+L_v+1)\times d_t}$ after dimensionality reduction can be expressed as:
\begin{equation}
    \overline{X'_l} = \mathrm{IDCT}(\mathrm{DCT}(X'_{l\ })_{[:, :d_t]})
\end{equation}
where $\mathrm{DCT}(\cdot)$ and $\mathrm{IDCT}(\cdot)$ represent discrete cosine transform and inverse cosine transform respectively. Then, we calculate the relationship between visual prompts-text tokens and text prompt-visual tokens, respectively. First, we use cosine similarity to calculate the affinity matrix:
\begin{align}
    A_{vt} &= \cos(\widetilde{P^{v}_l}, W_l^i)\in \mathbb{R}^{L_v\times N}\\
A_{tv} &= \cos(\widetilde{P^{t}_l}, \overline{x'_i}) \in \mathbb{R}^{L_t\times M}
\end{align}
where $\widetilde{P^{v}_l}$ and $\widetilde{P^{t}_l}$ represent the refined prompts obtained by self-structural prompt. Subsequently, we use the cross-modal structural relationship as a bridge to calculate the relationship between prompts based on the affinity matrix as follows:
\begin{align}
    A_{vv}& = \mathrm{exp}(-\beta\| A_{vt} - A_{vt}^\top \|^2_2)\\
A_{tt}& = \mathrm{exp}(-\beta\| A_{tv} - A_{tv}^\top \|^2_2)
\end{align}
where $\beta$ represents the scaling factor, which is usually fixed to 10. We then use a single layer of graph convolution to aggregate information for the prompts of each modality as follows:
\begin{align}
    \overline{P^v_l} &= \sigma(D^{-\frac{1}{2}}_{vv}A_{vv}D^{-\frac{1}{2}}_{vv}\widetilde{P^v_l}\theta_\mathrm{v1})\theta_\mathrm{v2}\\
\overline{P^t_l} &= \sigma(D^{-\frac{1}{2}}_{tt}A_{tt}D^{-\frac{1}{2}}_{tt}\widetilde{P^t_l}\theta_\mathrm{t1})\theta_\mathrm{t2}
\end{align}
where $\sigma(\cdot)$ is the activation function, $D$ represents the degree matrix and $\theta$ represents the learnable feature projection. Through the above operations, visual and textual prompts leverage the correlation of cross-modal tokens to convey information within the modality to ensure feature consistency while interacting information between modalities.

Finally, after obtaining the optimized prompts $\overline{P^{v}_l}$ and $\overline{P^{t}_l}$, we concatenate it with the original token sequence to obtain the refined token sequence $X'_l$ as the  input of the next layer, denoted as:
\begin{align}
    X'_{l} &= [c_l, x_l, \overline{P^{v}_l}]\\
W'_{l} &= [w_l, \overline{P^t_l}]
\end{align}

\subsection{Sample Probing}
\label{sec:sp}

In order to solve the overfitting phenomenon caused by simple samples, " i.e., samples that can be correctly classified by zero-shot CLIP with high confidence

Specifically, we define $p(y|x')$ as the zero-shot CLIP prediction probability for sample $x$, and $p(y|x)$ to represent the prediction probability after the prompt. The weight added by sample probing to the loss is expressed as follows:
\begin{equation}
    \alpha = \left (\frac{2\cdot |p(y|x') - p(y|x)|}{|p(y|x') + p(y|x)|}\right )^\gamma
\end{equation}
where $\gamma$ is the scaling hyperparameter. We can find from the formula that only when the probability of the zero-shot CLIP and the predicted logits after the prompt are high in the label class and the gap is small, the weight $\alpha$ becomes lower. Otherwise, the loss weight of the sample will be increased. This measurement method can avoid the inaccuracy problem caused by inferring a single prediction result in previous work \cite{amutuning}. Finally, we apply $\max(\alpha, 1)$ to clip the loss weight to avoid extreme values. It is worth noting that the weights obtained by our sample probing are only used for loss calculation, and we still use $p(y|x)$ for inference.

\subsection{Loss Calculation}
\label{sec:loss}
For sample $x$, the visual and text features obtained after the pre-trained backbone and Self-Structural and Cross-Structural prompts are denoted as $x$, $w$, respectively, and the predicted probability of the corresponding ground truth label $y$ is $p(y|x)$. At the same time, the visual and text features extracted by zero-shot CLIP are denoted as $x'$ and $w'$, respectively. We first apply the standard cross-entropy loss and scale the cross-entropy loss through our proposed sample probing. Secondly, we refer to \cite{kgcoop, lamm}, and use the frozen CLIP features to regularize the prompted features. Our loss calculation can be as follows:
\begin{align}
        \mathcal{L}_{\mathrm{reg}} &= \omega_v(1-\cos(x', x))+\omega_t(1-\cos(w', w))\\
        \mathcal{L} &= (1+\alpha)\mathcal{L}_{\mathrm{ce}}(y, p(y|x)) + \mathcal{L}_{\mathrm{reg}}
\end{align}
where $\omega_v$ and $\omega_t$ represent the hyperparameters of the regularization weights of visual features and text features, respectively.

\begin{table*}
    \caption{Comparison with SOTA methods on different datasets using 16 samples in the Base-to-New generalization setting. “Base” and “New” are the recognition accuracies on base and novel classes respectively. “HM” is the harmonic mean of base and new accuracy, that demonstrates the trade-off between adaption and generalization.}
    \label{tab:main}
    \centering
    \setlength{\tabcolsep}{2.5pt}
    \begin{tabular}{rccc|ccc|ccc|ccc} 
    
    \toprule
         \multirow{2}{*}{Methods} &  \multicolumn{3}{c|}{Average}&  \multicolumn{3}{c|}{ImageNet}&  \multicolumn{3}{c|}{Caltech101}&  \multicolumn{3}{c}{OxfordPets}\\ 
          &Base & New & HM& Base & New & HM& Base & New & HM&Base & New & HM\\
\midrule
         CLIP \cite{clip}&  69.34& 74.22&71.70&  72.43& 68.14&70.22&  96.84& 94.00&95.40&  91.17& 97.26&94.12
\\ 
         CoOp \cite{coop}&   82.69& 63.22&71.66&   76.47& 67.88&71.92&   98.00& 89.81&93.73&   93.67& 95.29&94.47
\\ 
         CoCoOp \cite{cocoop}&   80.47& 71.69&75.83&   75.98& 70.43&73.10&   97.96& 93.81&95.84&   95.20& 97.69&96.43
\\ 
         ProDA\cite{proda}& 81.56& 72.30& 76.65& 75.40& 70.23& 72.72& 98.27& 93.23& 95.68& \textbf{95.43}& 97.83& 96.62 \\
         KgCoOp \cite{kgcoop}&   80.73& 73.60&77.00&   75.83& 69.96&72.78&   97.72& 94.39&96.03&   94.65& 97.76&96.18 \\
         MaPLe \cite{maple}&   82.28& 75.14&78.55&   76.66& 70.54&73.47&   97.74& 94.36&96.02&   \textbf{95.43}& 97.76&96.58
\\ 
 TCP \cite{tcp}& 84.13& 75.36& 79.51& 77.27& 69.87& 73.38& 98.23&\textbf{ 94.67}& 96.42& 94.67& 97.20&95.92\\
MMA \cite{mma}& 83.20& \textbf{76.80}& 79.87& 77.31& \textbf{71.00}& \textbf{74.02}& 98.40& 94.00& 96.15& 95.40& \textbf{98.07}&\textbf{96.72}
\\
 \midrule
 ISP (Ours) & \textbf{85.72}& 76.24&\textbf{80.70}& \textbf{77.50}&70.40 & 73.78& \textbf{98.64}& 94.31&\textbf{96.43}& 95.29& 97.44& 96.35\\
 \midrule
 \midrule
         \multirow{2}{*}{Methods} &  \multicolumn{3}{c|}{StanfordCars}&  \multicolumn{3}{c|}{Flowers102}&  \multicolumn{3}{c|}{Food101}&  \multicolumn{3}{c}{FGCVAircraft}\\ 
          &Base & New & HM& Base & New & HM& Base & New & HM&Base & New & HM\\
          \midrule
         CLIP \cite{clip}
&  63.37& 74.89&68.65&  72.08& \textbf{77.80}&74.83&  90.10& 91.22&90.66&  27.19& 36.29&31.09
\\ 
         CoOp \cite{coop}
&   78.12& 60.40&68.13&   97.60& 59.67&74.06&   88.33& 82.26&85.19&   40.44& 22.30&28.75
\\ 
         CoCoOp \cite{cocoop}
&   70.49& 73.59&72.01&   94.87& 71.75&81.71&   90.70& 91.29&90.99&   33.41& 23.71&27.74
\\ 
         ProDA\cite{proda}& 74.70& 71.20& 72.91& 97.70& 68.68&80.66 & 90.30& 88.57& 89.43& 36.90& 34.13& 35.46 \\
         KgCoOp \cite{kgcoop}& 71.76& 75.04& 73.36& 95.00& 74.73&83.65 & 90.50& 91.70&91.09&   36.21& 33.55&34.83 \\
         MaPLe \cite{maple}
&   72.94& 74.00&73.47&   95.92& 72.46&82.56&   \textbf{90.7}1& \textbf{92.05}&\textbf{91.38}&   37.44& 35.61&36.50
\\ 
 TCP \cite{tcp}
& 80.80& 74.13& 77.32& 97.73& 75.57& 85.23& 90.57& 91.37& 90.97&41.97 &34.43 &37.83\\
MMA \cite{mma}
& 78.50& 73.10& 75.70& 97.77& 75.93& 85.48& 90.13& 91.30& 90.71& 40.57& \textbf{36.33}&38.33
\\
 \midrule
 ISP (Ours) & \textbf{83.38}&\textbf{75.20} &\textbf{79.08}& \textbf{98.58}& 75.74& \textbf{85.67}& 90.27&91.53&90.89&\textbf{48.40}&34.73&\textbf{40.44}\\
 \midrule
 \midrule
         \multirow{2}{*}{Methods} &  \multicolumn{3}{c}{SUN397}&  \multicolumn{3}{c|}{DTD}&  \multicolumn{3}{c|}{EuroSAT}&  \multicolumn{3}{c}{UCF101}\\ 
          &Base & New & HM& Base & New & HM& Base & New & HM&Base & New & HM\\
 \midrule
         CLIP \cite{clip}
&  69.36& 75.35&72.23&  53.24& 59.90&56.37&  56.48& 64.05&60.03&  70.53& 77.50&73.85
\\ 
         CoOp \cite{coop}
&   80.60& 65.89&72.51&   79.44& 41.18&54.24&   92.19& 54.74&68.69&   84.69& 56.05&67.46
\\ 
         CoCoOp \cite{cocoop}
&   79.74& 76.86&78.27&   77.01& 56.00&64.85&   87.49& 60.04&71.21&   82.33& 73.45&77.64
\\ 
 ProDA\cite{proda}& 78.67& 76.93& 77.79& 80.67& 56.48& 66.44& 83.90& 66.00& 73.88& 85.23& 71.97&78.04 \\
 KgCoOp \cite{kgcoop}&   80.29& 76.53&78.36&   77.55&54.99&64.35& 85.64& 64.34& 73.48& 82.89& 76.67&79.65 \\
         MaPLe \cite{maple}
&   80.82& 78.70&79.75&   80.36& 59.18&68.16&   94.07& 73.23&82.35&   83.00& 78.66&80.77
\\ 
 TCP \cite{tcp}
& \textbf{82.63}&78.20 &80.35 &82.77 &58.07 &68.25 & 91.63& 74.73& 82.32&87.13 &80.77 &83.83\\
MMA \cite{mma}
& 82.27& 78.57& 80.38& 83.20& \textbf{65.63}& \textbf{73.38}& 85.46& \textbf{82.34}& 83.87& 86.23& 80.03&82.20
\\
 \midrule
 ISP (Ours) & 82.33& \textbf{79.02}&\textbf{80.64}&\textbf{85.07 }&61.06&71.09&\textbf{95.87}&78.03&\textbf{86.04}& \textbf{87.61}&\textbf{81.18}&\textbf{84.21}\\
\bottomrule
    \end{tabular}

\end{table*}

\section{Experiment}
\label{sec:experiment}

\subsection{Experimental Settings}
\label{sec:experimentsetting}
\subsubsection{Benchmark Settings}
\paragraph{Base-New Generalization}
To evaluate the generalization performance of our work, following previous works \cite{maple, cocoop, coop}, we adopt the setting of few-shot learning in 11 classification datasets. The datasets are divided into two non-overlapping subsets of base and new classes by category. The model is fine-tuned on the base class with only 16 samples in each class, and the obtained model is tested on the new class directly.
Like many previous works, these 11 classification datasets include two general object recognition datasets: ImageNet \cite{imagenet} and Caltech101 \cite{caltech101}; five fine-grained visual classification datasets: Oxford Pets \cite{oxfordpet}, Stanford Cars \cite{oxfordpet}, Flowers 102 \cite{flower}, Food 101 \cite{food}, and FGVC Aircraft \cite{fgvcaircraft}; scene understanding dataset SUN397 \cite{sun397}; material dataset DTD \cite{dtd}; satellite image recognition dataset EuroSAT \cite{eurosat}; and action classification dataset UCF101 \cite{ucf101}. These datasets contain large-scale recognition tasks, which can evaluate the generalization ability of the model comprehensively.

\paragraph{Cross-dataset Evaluation}

We evaluated the transfer ability of our work on cross-datasets. In line with CoOp \cite{coop}, we first train the model on 1000 classes of the ImageNet dataset with 16 samples per class. The trained model is then directly applied to the rest of the above-mentioned datasets for testing. Unlike Base-New Generalization, samples are sampled from all classes in the dataset during training and testing.

\paragraph{Domain Generalization}

To verify the performance of our method on Out-of-Distribution datasets, Zhu et al. \cite{cocoop} proposed fine-tuning the model on ImageNet \cite{imagenet} and testing it on several other variants of the domain shift mode. These variants are ImageNet-V2 \cite{imv2}, ImageNet-Sketch \cite{im-s}, ImageNet-A \cite{im-a}, and ImageNet-R \cite{im-r}, respectively.

\subsubsection{Implementation Details}

Following previous works \cite{coop, maple}, we conduct the experiments under the few-shot setting with 16 labeled samples per class. We use the CLIP model pre-trained on the ViT-B/16 as the backbone network for all experiments. In the base-new generalization setting, we apply our proposed ISP from layer 1 to layer 12. We adopt randomly initialized vectors from the Gaussian distribution as initial values for text and visual prompts. The lengths of text prompt $L_t$ and visual prompts $L_v$ are set to 6 and 4, respectively. We train our model for 5 epochs on ImageNet and SUN397 datasets and 50 epochs on other datasets with a batch size of 16. For cross-dataset and domain generalization settings, we apply our proposed ISP from layers 10 to 12 and train the model for 2 epochs. The loss scale coefficient $\gamma$ in our proposed sample probing is set to 0.3. We adopt the Adam optimizer and apply a cosine-annealing learning rate scheduling strategy with an initial learning rate of 0.025. All the experiments are conducted on a single NVIDIA RTX 4090. Our experiments adopt Pytorch as the framework and use mixed precision settings to accelerate the experiments.
We report the average classification accuracy of the base and novel classes and their harmonic means under three seeds (1/2/3). 

\subsection{Comparison with SOTA Methods}
\label{sec:sota}
\subsubsection{Base-New Generalization}
We first compare our proposed ISP with existing SOTA methods, including the baseline method CLIP, the methods enhance the text branch, CoOp \cite{coop}, CoOpOp \cite{cocoop}, KgCoOp \cite{kgcoop}, TCP \cite{tcp}, and the methods designed with multi-modal interaction, i.e., MMA \cite{mma}, MaPLE \cite{maple}. We report the base class and new class accuracies and their harmonic means (HM) on 11 widely used benchmarks in Table \ref{tab:main}. From the results, we can observe that our work achieves the highest results in the average base class accuracy and average harmonic mean accuracy on 11 datasets. At the same time, compared with the state-of-the-art methods, our work achieves higher results in the harmonic mean accuracy in 7/11 datasets. These results demonstrate a strong generalization ability when facing new classes of ISP. Compare with existing text brach-based methods such as CoOp, CoOpOp, KgCoOp, and TCP, 
Our method learns the features of the visual branch and text branch at the same time, thereby obtaining better results with an average HM of 80.70\%. MaPLe applies prompts to each layer of the visual and text encoders and establishes a one-way information transfer from text modality prompts to visual modality prompts. Our method simultaneously refines the prompts within each modality, builds information transfer between two modalities, and achieves improvements of 3.44\%, 1.00\%, and 2.15\% on base class accuracy, new class accuracy, and harmonic mean, respectively. 
MMA adds shared feature projection layers in the visual and text branches to exchange information between modalities. Our work uses cross-modal structural information to guide prompt features and achieves a 0.95 improvement on HM. 
These results demonstrate the superiority of our ISP over SOTA methods.
\begin{table*}[ht]

    \centering
    \caption{Comparison Results in Cross-dataset Evaluation Setting}
    \label{tab:cross}
    \begin{adjustbox}{width = \linewidth}
    \setlength{\tabcolsep}{4pt}
    \begin{tabular}{rc|cccccccccc|c}
    \toprule
         \multirow{2}{*}{Methods}& Source & \multicolumn{11}{c}{Target}\\
         \cmidrule{2-13}
         &  ImageNet&  Caltech&  Pets&  Cars&  Flowers&  Food&  Aircraft&  SUN&  DTD& EuroSAT& UCF101&Average\\
         \midrule
         CoOp \cite{coop}&  \textbf{71.51}&  93.70&  89.14&  64.51&  68.71&  85.30&  18.47&  64.15&  41.92& 46.39& 66.55&63.88
\\
         CoCoOp \cite{cocoop}&  71.02&  \textbf{94.43}&  90.14&  65.32&  71.88&  86.06&  22.94&  67.36&  45.73& 45.37& 68.21&65.74\\
 MaPLe \cite{maple}& 70.72& 93.53& 90.49& 65.57& 72.23& 86.20& 24.74& 67.01& 46.49& 48.06& 68.69&66.30
\\
 TCP \cite{tcp}& 71.40& 93.97& \textbf{91.25}& 64.69& 71.21& \textbf{86.69}& 23.45& 67.15& 44.35& \textbf{51.45}& 68.73&66.29\\
 MMA \cite{mma}& 71.00& 93.80& 90.30& 66.13& 72.07& 86.12& \textbf{25.33}& \textbf{68.17}& 46.57& 49.24& 68.32&66.61
\\
 \midrule
 ISP (Ours)& 71.10& 93.37& 90.32& \textbf{66.72}& \textbf{72.30}& 86.32& 24.88& 67.19& \textbf{47.30}& 49.54& \textbf{69.59}&\textbf{66.75}\\
 \bottomrule
    \end{tabular}
\end{adjustbox}
\end{table*}

\subsubsection{Cross-Dataset Evaluation}

We compare the results of our work with existing SOTA methods in the Cross-dataset Evaluation setting in Table \ref{tab:cross}. Compared with the existing methods, the proposed ISP achieved the highest average accuracy of 66.75\%. The performance of our work on the ImageNet dataset surpassed MMA and MaPLe. In the target datasets, our work achieves SOTA accuracy in 4/10 datasets. To be specific, our method outperforms CoOp in 9/10 datasets, Maple in 8/10 datasets, MMA in 7/10 datasets, and TCP in 6/10 datasets. These results demonstrate the superior ability of our method to generalize downstream tasks.
\begin{table}[ht]
    \centering
        \caption{Comparison Results in Domain Generalization Setting. "-V2", "-A", "-S", "-R" represent the four out of distribution variants of ImageNet, namely ImageNet-V2, ImageNet-adversarial, ImageNet-Sketch, ImageNet-Rendition.}
    \label{tab:ood}
    \begin{tabular}{rccccc}
    \toprule
         Methods&  ImageNet&  -V2&  -S&  -A& -R\\
         \midrule
         CLIP \cite{clip}&  66.73&  60.83&  46.15&  47.77& 73.96\\
         CoOp \cite{coop}&  \textbf{71.51}&  64.20&  47.99&  49.71& 75.21\\
         CoCoOp \cite{cocoop}&  71.02&  64.07&  48.75&  50.63& 76.18\\
 KgCoOp \cite{kgcoop}& 71.20& 64.10& 48.97& 50.69&76.70\\
         MaPLe \cite{maple}&  70.72&  64.07&  49.15&  50.90& 76.98\\
         MMA \cite{mma}&  71.00&  64.33&  49.13&  \textbf{51.12}& 77.32\\
         \midrule
         ISP&  71.10&  \textbf{64.50}&  \textbf{49.40}&  49.90& \textbf{77.40}\\
         \bottomrule
    \end{tabular}

\end{table}
\subsubsection{Domain Generalization}
We validated our work on the setting of training on ImageNet and directly tested the Domain Generalization capability in four out-of-distribution (OOD) datasets. The results are shown in Table \ref{tab:ood}. From the results, we can see that our work has achieved competitive performance. Specifically, our work achieves competitive results in the ImageNet dataset and outperforms SOTA methods on 3/4 of the OOD datasets. These results demonstrate the robustness of our work in the domain generalization setting.

\subsection{Ablation Studies}
\label{sec:ablation}
\begin{table}[ht]
    \centering
        \caption{Results of the Component Ablation of the proposed method. 
        }
    \label{tab:ablation}
    \begin{tabular}{lccccc}
    \toprule
          \multicolumn{3}{c}{Componet}&  \multicolumn{3}{c}{Average Accuracy (\%)}\\
          \cmidrule{1-6}
          SP &SS&  CS&  Base& New&H Mean\\
         \midrule
  -&-& -&83.15& 74.00&78.31\\
    \checkmark&-& -&  83.49& 74.55&78.77\\
          -&\checkmark&  -&  85.63& 75.41&80.20\\
          -&\checkmark&  \checkmark&  85.67& 75.65&80.35\\
  \checkmark&\checkmark& -& 85.62& 75.77&80.39\\
  \checkmark&\checkmark& \checkmark&\textbf{85.72}& \textbf{76.24}&\textbf{80.70}\\
 \bottomrule
    \end{tabular}

\end{table}
\subsubsection{Component Ablation}
We verify the effectiveness of the proposed three components, i.e., self-structural prompt, cross-structural prompt, and sample probing, respectively by conducting ablation experiments. From Table \ref{tab:ablation}, when only SP is applied, the base class accuracy, new class accuracy, and HM accuracy are improved simultaneously, which proves that SP can effectively improve the discrimination ability of the network and will not cause the new class performance to decrease due to the improvement of the base class accuracy. When SS is applied, the performance of both the base class and the new class is significantly improved, which proves that optimizing the prompts by utilizing the relationship between the prompts and the tokens can improve the generalization ability of the model. After adding SP, the performance of HM is improved, which demonstrates the adaptability of the proposed SP. When SS is combined with CS, the accuracy of the base class and the new class is further improved because the cross-modal information is used to optimize the prompts further. When the three parts are enabled at the same time, the model achieves the highest performance, especially when the new class accuracy is greatly improved. The above results prove the effectiveness of our proposed module in few-shot learning tasks.

\begin{table}[h]
    \centering
    \caption{Ablation Experiment of the Length of Text and Visual Prompts}
    \label{tab:length}
    \begin{tabular}{cccccc}
    \toprule
        $L_t$ ($L_v=4$)&  2&  4&  6&  8& 10\\
         \midrule
         HM (\%) &  76.57  &  79.85&  80.70&  80.12& 80.16\\
         \midrule
         $L_v$ ($L_t=6$)&  2&  4&  6&  8& 10\\
          \midrule
 HM (\%)& 80.23& 80.70& 80.15& 80.50&80.49\\
 \bottomrule
    \end{tabular}

\end{table}
\subsubsection{Lengths of Visual and Text Prompts}
Since our proposed work exploits the structural relationship between prompts and visual, textual tokens, we conducted ablation experiments on the lengths of visual and text prompts, as shown in Table \ref{tab:length}. From the results, we can observe that when the number of prompts gradually increases, the performance first gradually improves and then slightly decreases. In particular, when the number of textual prompts is too small, the network performance decreases significantly. This is because fewer textual prompts cannot learn domain-related information well and may overfit the limited category information.
Finally, we chose the optimal combination, i.e., the length of text prompts and visual prompts in this paper are 6 and 4, respectively.


\begin{table}[h]
    \caption{Results for different values of the scale factor $\gamma$}
    \label{tab:saclefactor}
    \centering
    \begin{tabular}{ccccclcc}
    \toprule
         $\gamma$&  0.1&  0.2&  0.3 &  0.5 &0.7&  1.0& 2.0 \\
         \midrule
 HM (\%)& 80.44& 80.57& 80.70 & 80.61 &80.46& 79.95& 80.20\\
 \bottomrule
    \end{tabular}

\end{table}

\subsubsection{Scaling factor for Sample Probing}
When calculating the loss coefficient in the sample probing module, we used the hyperparameter $\gamma$ to control the scaling factor of different difficulties. We verified the impact of different values on the results through experiments. The results are shown in Table \ref{tab:saclefactor}, 
when $\gamma$ is larger, sample probing tends to reduce the loss of simple samples. However, a larger $\gamma$ will greatly reduce the loss of easily confused samples. The experimental results show that the network achieves the best performance when $\gamma$ is 0.3 and only the samples with the is applied with lower loss scale coefficient. This result shows that adding less loss to simple samples can bring better results and reduce overfitting. 

\begin{figure}[!ht]
    \centering
    \includegraphics[width=0.75\linewidth]{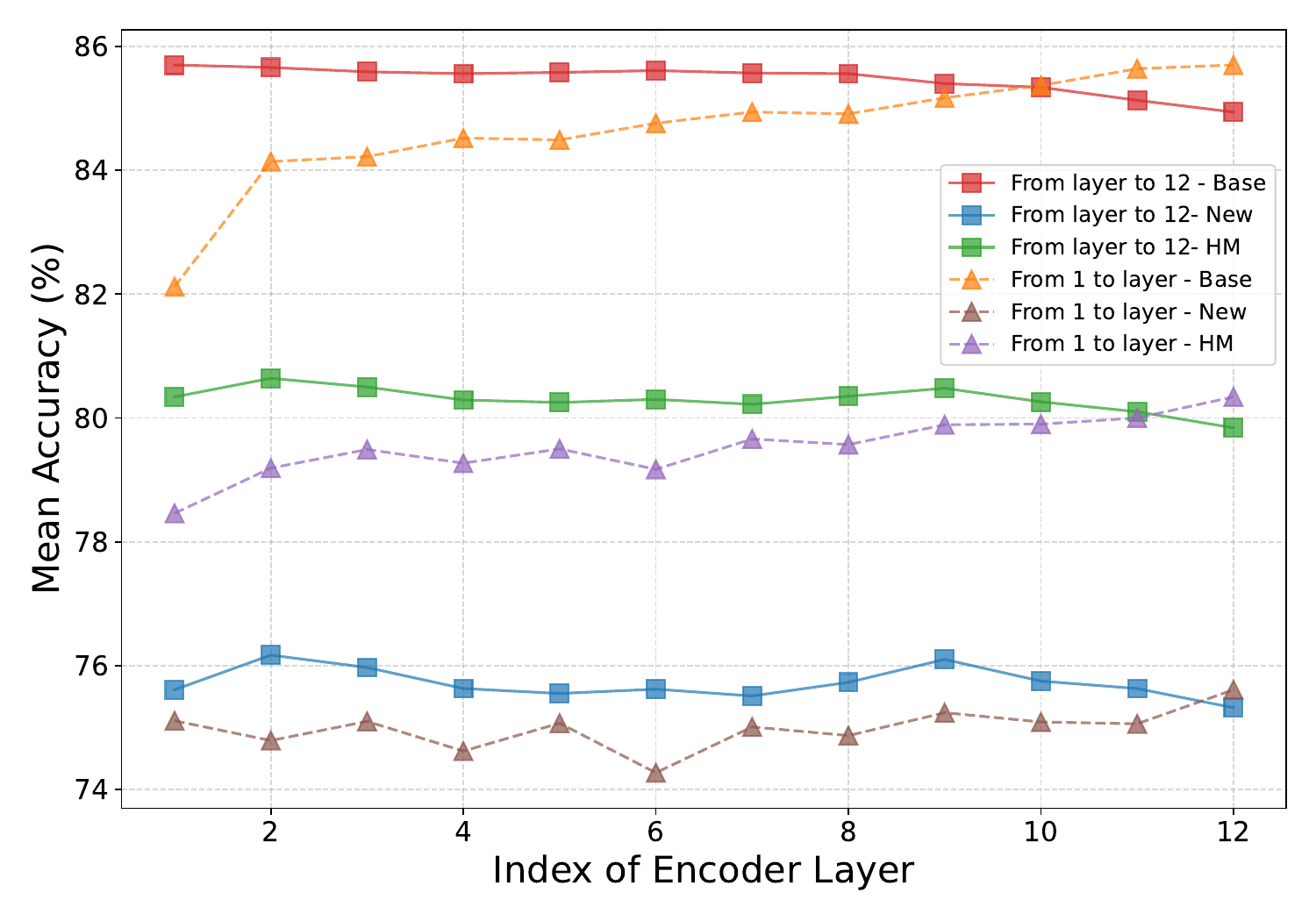}
    \caption{Results of different adding layers of self-structural prompt and cross-structural prompt.}
    \label{fig:differentlayer}
\end{figure}

\subsubsection{Layers of Adding SSP and CSP}
Our proposed self-structural prompt (SSP) and cross-structural prompt (CSP) are added after the frozen visual and text encoder layers. We validated different numbers and ways of adding SSP and CSP by experiments and reported the average accuracy, as shown in Figure \ref{fig:differentlayer}. From the results, we can observe that when the number of added layers gradually increases, the average accuracy of the base class increases linearly, which demonstrates that our proposed method can achieve consistent improvement on the base class with more parameters. In terms of new class accuracy, we find that gradually adding SSP and CSP  from 2 to 6 layers leads to a decrease in accuracy, while adding after the 10th layer will bring an improvement. 
The above results prove that our work does not suffer from serious overfitting due to a large number of parameters and has higher flexibility.
\subsection{Visualization Analysis}
\subsubsection{Image Feature Representation Analysis}
\begin{figure}[t]
    \centering
    \includegraphics[width=0.75\linewidth]{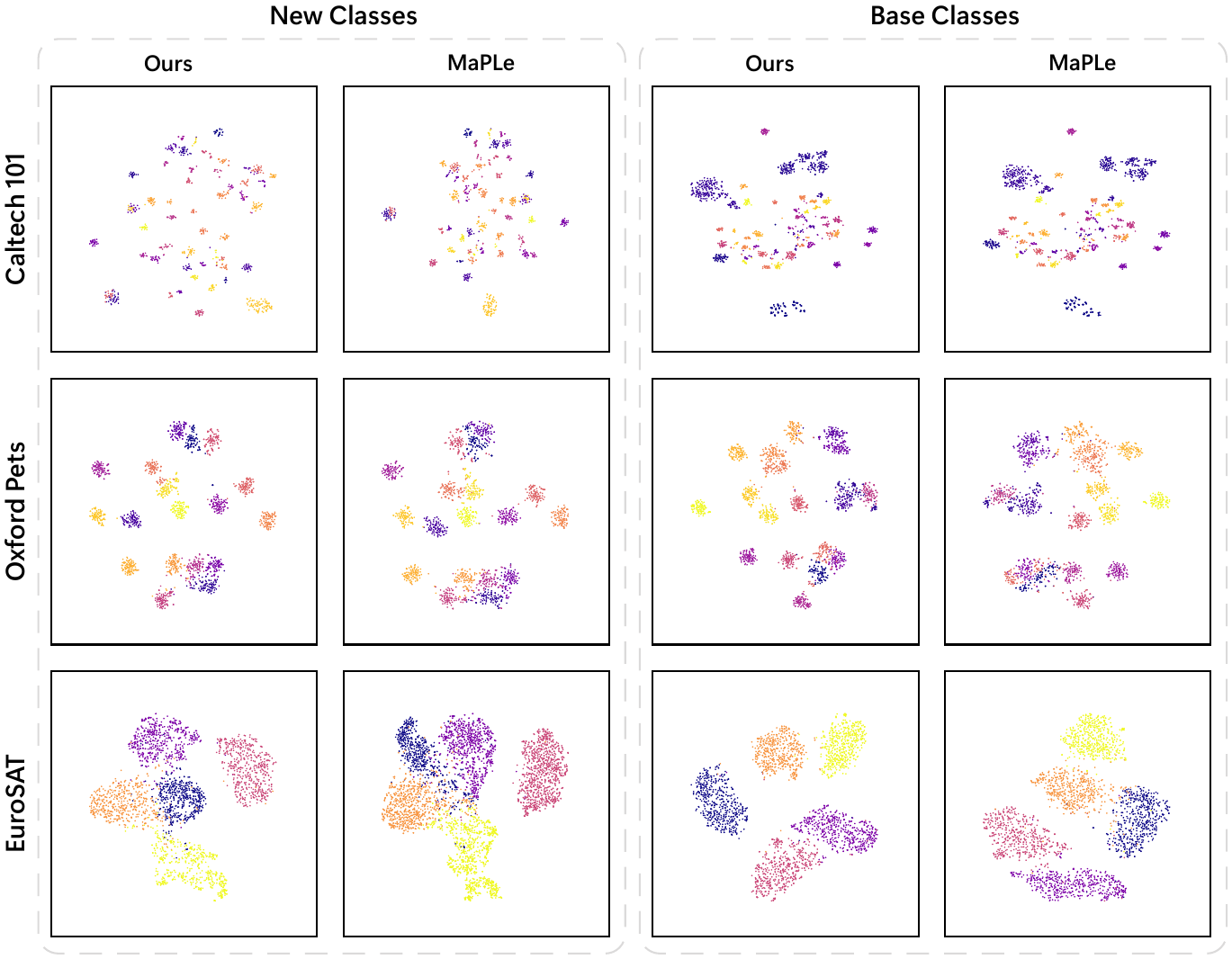}
    \caption{T-SNE visualization of image features by adopting our method and MaPLe on three types of datasets. Each point represents a sample and the color indicates its category.}
    \label{fig:tsne}
\end{figure}

\begin{figure}[t]
    \centering
    \includegraphics[width=0.75\linewidth]{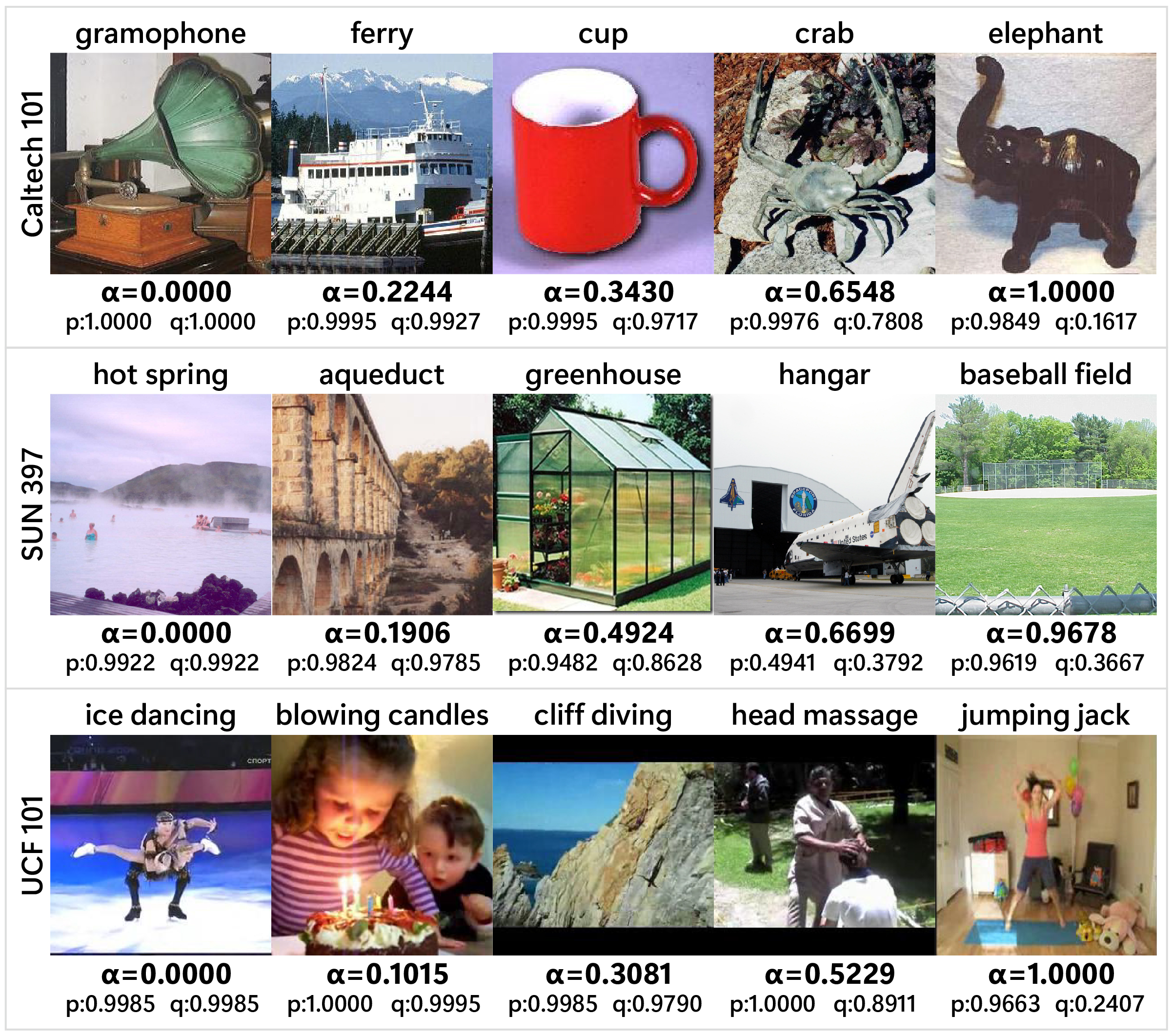}
    \caption{Loss coefficient of samples in different datasets calculated by the proposed sample probing.  $p$ represents the prediction probability after prompt, and $q$ represents the prediction  probability of zero-shot CLIP.}
    \label{fig:visualsp}
\end{figure}

\label{sec:visual}

To verify the effectiveness of our work more intuitively, we select one from each of three types of datasets (general object recognition datasets, fine-grained visual classification datasets, and specialized datasets) to analyze the image feature representation and visualize them through t-SNE, as shown in Figure \ref{fig:tsne}.  Compared with MaPLe \cite{maple}, which also uses multi-modal information, our work demonstrates better class separability and features are more compact within each class in both new classes and base classes. We attribute the above results to the utilization of the structural relationship between prompts and tokens, which further improves the discriminability of pre-trained image feature representation.  Although image features need to match text features in classification, image features that are more compact within classes and more separated between classes can be more effectively aligned with text class features, resulting in more accurate classification.  At the same time, text features are also refined by the self-structural relationships and the cross-structural relationships with image features. Under the guidance of the refined text features, the network distinguishes the categories of input image features better.

\subsubsection{ Loss Coefficient of Samples}

We visualize the loss coefficients generated by the proposed sample probing, as shown in Figure \ref{fig:visualsp}. In the figure, $p $ represents the predicted probability after the prompt, while $q $ denotes the predicted probability of zero-shot CLIP. For the visualization, we randomly select samples from five intervals of $\alpha $ values. From the results, we derive the following observations:  
1) After the prompt learning, the network classifies samples with higher probability, whereas the predicted probabilities of zero-shot CLIP exhibit significant variation. The low prediction probability of CLIP may arise from the challenge in distinguishing background and foreground objects (e.g., crab), or from the large intra-class variance in samples of the same category (e.g., elephant), as well as the relatively uncommon class names (e.g., basset hound).  
2) When zero-shot CLIP already performs well in classification and prompt learning does not lead to further improvement, the proposed sample probing does not introduce additional loss to the sample.  
3) A large gap between $p $ and $q $ indicates that the sample is difficult, and we assign it a higher loss coefficient.  

\section{Conclusion}
\label{sec:conclusion}
This paper presents an integrated structural prompt method to address the challenges of transferring vision-language models in downstream tasks. Through self-structural prompt and cross-structural prompt modules, we fully exploit the structural relationships between prompts and text/visual tokens and enhance information transfer within and across modalities. Furthermore, the sample probing module dynamically adjusts loss coefficients based on sample difficulty, effectively reducing overfitting to simple samples and improving the generalization ability of the model to new classes. Experimental results show that our method achieves leading performance on multiple datasets, particularly excelling in new class generalization, cross-dataset evaluation, and domain generalization settings. Future work may further explore optimizing prompt designs across different tasks and datasets to enhance the generalization capabilities of VLMs. 
%
%
%
\bibliographystyle{splncs04}
\bibliography{ISP}
%

	\Acknowledgements{This work was supported by the National Natural Science Foundation of China under Grant 61860206004, by the Natural Science Foundation of Anhui Province under Grant 2108085Y23, by the Natural Science Foundation for the Higher Education Institutions of Anhui Province under Grant KJ2021A0038, and by the University Synergy Innovation Program of Anhui Province under Grant GXXT-2022-032 and by Anhui Provincial Key Research and Development Program under Grant
2022i01020014. }

	
%
%
%


	
\end{document}